\documentclass[10pt,logo,copyright]{nvidiatechreport}
\usepackage[authoryear,sort&compress,round]{natbib}

\usepackage[utf8]{inputenc} 
\usepackage[T1]{fontenc}    

\usepackage{parskip}        
\usepackage{url}            
\usepackage{booktabs}       
\usepackage{amsfonts}       
\usepackage{nicefrac}       
\usepackage{microtype}      
\usepackage{xcolor}         
\usepackage[dvipsnames]{xcolor} 
\usepackage{graphicx}
\usepackage{animate}        
\usepackage{subcaption}
\usepackage{tabularx}
\usepackage{makecell}
\usepackage{adjustbox}
\usepackage{setspace}
\newcolumntype{M}[1]{>{\centering\arraybackslash}m{#1}}
\usepackage{float}
\usepackage{tikz}
\usetikzlibrary{positioning,shapes,arrows}
\usepackage{amsmath,amsfonts,bm, bbm,leftindex}
\usepackage{multirow}
\usepackage{comment}
\usepackage{gensymb}
\usepackage{lipsum}
\usetikzlibrary{arrows.meta, positioning, fit}
\usepackage[para]{threeparttable}
\usepackage{tikz}
\usetikzlibrary{tikzmark}
\usepackage{microtype}
\usepackage{graphicx}
\usepackage{tabularx}
\usepackage{booktabs}
\usepackage{arydshln}
\usepackage{amsmath}
\usepackage{amssymb}
\usepackage{mathtools}
\usepackage{amsthm}
\usepackage{longtable}
\usepackage{booktabs}
\usepackage{array}
\usepackage{xspace}
\usepackage{wrapfig}
\usepackage{tcolorbox}
\usepackage{listings}
\tcbuselibrary{listings}

\lstdefinelanguage{json}{
    basicstyle=\small\ttfamily,
    showstringspaces=false,
    breaklines=true,
    stringstyle=\color{blue},
    keywordstyle=\color{purple},
    commentstyle=\color{gray},
    morestring=[b]",
}
\newtcolorbox{sectionbox}[2][]{
  colback=gray!5,
  colframe=gray!75!black,
  fonttitle=\bfseries,
  title=#2,
  #1
}
\newtcblisting{codebox}[1][]{
    listing only,
    listing options={
        basicstyle=\small\ttfamily,
        breaklines=true,
        showstringspaces=false,
        #1
    },
    colback=gray!10,
    colframe=gray!40,
    arc=4mm,
    boxrule=0.5pt,
    left=2mm,
    right=2mm,
    top=0mm,
    bottom=0mm,
}

\theoremstyle{plain}

\theoremstyle{definition}

\theoremstyle{remark}

\usepackage[capitalize,noabbrev]{cleveref}

\usepackage{tcolorbox}
\usepackage{listings}
\tcbuselibrary{listings}

\def\eqref#1{equation~\ref{#1}}

\def\1{\bm{1}}

\DeclareMathAlphabet{\mathsfit}{\encodingdefault}{\sfdefault}{m}{sl}
\SetMathAlphabet{\mathsfit}{bold}{\encodingdefault}{\sfdefault}{bx}{n}

\makeatletter
\let\save@mathaccent\mathaccent
\newcommand*\if@single[3]{%
  \setbox0\hbox{${\mathaccent"0362{#1}}^H$}%
  \setbox2\hbox{${\mathaccent"0362{\kern0pt#1}}^H$}%
  \ifdim\ht0=\ht2 #3\else #2\fi
  }
\newcommand*\rel@kern[1]{\kern#1\dimexpr\macc@kerna}
\newcommand*\widebar[1]{\@ifnextchar^{{\wide@bar{#1}{0}}}{\wide@bar{#1}{1}}}
\newcommand*\wide@bar[2]{\if@single{#1}{\wide@bar@{#1}{#2}{1}}{\wide@bar@{#1}{#2}{2}}}
\newcommand*\wide@bar@[3]{%
  \begingroup
  \def\mathaccent##1##2{%
    \let\mathaccent\save@mathaccent
    \if#32 \let\macc@nucleus\first@char \fi
    \setbox\z@\hbox{$\macc@style{\macc@nucleus}_{}$}%
    \setbox\tw@\hbox{$\macc@style{\macc@nucleus}{}_{}$}%
    \dimen@\wd\tw@
    \advance\dimen@-\wd\z@
    \divide\dimen@ 3
    \@tempdima\wd\tw@
    \advance\@tempdima-\scriptspace
    \divide\@tempdima 10
    \advance\dimen@-\@tempdima
    \ifdim\dimen@>\z@ \dimen@0pt\fi
    \rel@kern{0.6}\kern-\dimen@
    \if#31
      \overline{\rel@kern{-0.6}\kern\dimen@\macc@nucleus\rel@kern{0.4}\kern\dimen@}%
      \advance\dimen@0.4\dimexpr\macc@kerna
      \let\final@kern#2%
      \ifdim\dimen@<\z@ \let\final@kern1\fi
      \if\final@kern1 \kern-\dimen@\fi
    \else
      \overline{\rel@kern{-0.6}\kern\dimen@#1}%
    \fi
  }%
  \macc@depth\@ne
  \let\math@bgroup\@empty \let\math@egroup\macc@set@skewchar
  \mathsurround\z@ \frozen@everymath{\mathgroup\macc@group\relax}%
  \macc@set@skewchar\relax
  \let\mathaccentV\macc@nested@a
  \if#31
    \macc@nested@a\relax111{#1}%
  \else
    \def\gobble@till@marker##1\endmarker{}%
    \futurelet\first@char\gobble@till@marker#1\endmarker
    \ifcat\noexpand\first@char A\else
      \def\first@char{}%
    \fi
    \macc@nested@a\relax111{\first@char}%
  \fi
  \endgroup
}
\makeatother

\definecolor{darkred}{rgb}{0.7, 0.0, 0.0}

\usepackage{pifont}
\newcommand{\ourmethod}{Terminal-Task-Gen\xspace}
\newcommand{\ourdata}{Terminal-Corpus\xspace}
\newcommand{\ourmodel}{Nemotron-Terminal\xspace}

\newcommand{\crefnames}[3]{%
  \@for\next:=#1\do{%
    \expandafter\crefname\expandafter{\next}{#2}{#3}%
  }%
}

\title{On Data Engineering for Scaling LLM Terminal Capabilities}

\author{\vspace{-.3cm}
Renjie Pi\footnote[1]{Equal technical contribution.  Correspondence to: <renjiep@nvidia.com>, <gralam@nvidia.com>, <wping@nvidia.com>},~Grace Lam$^*$,~Mohammad~Shoeybi, ~Pooya Jannaty, ~Bryan~Catanzaro,~
Wei Ping\footnote[2]{Leads the effort.}
}
\begin{abstract}
  Despite rapid recent progress in the terminal capabilities of large language models, the training data strategies behind state-of-the-art terminal agents remain largely undisclosed. We address this gap through a systematic study of data engineering practices for terminal agents, making two key contributions: (1) \textbf{\ourmethod}, a lightweight synthetic task generation pipeline that supports seed-based and skill-based task construction, and (2) a comprehensive analysis of data and training strategies, including filtering, curriculum learning, long context training, and scaling behavior. Our pipeline yields \textbf{\ourdata}, a large-scale open-source dataset for terminal tasks. Using this dataset, we train \textbf{\ourmodel}, a family of models initialized from Qwen3(8B, 14B, 32B) that achieve substantial gains on Terminal-Bench 2.0: \ourmodel-8B improves from 2.5\% to 13.0\% \ourmodel-14B improves from 4.0\% to 20.2\%, and \ourmodel-32B improves from 3.4\% to 27.4\%, matching the performance of significantly larger models. To accelerate research in this domain, we open-source our model checkpoints and most of our synthetic datasets at \url{https://huggingface.co/collections/nvidia/nemotron-terminal}.
\end{abstract}

\begin{document}

\maketitle

\abscontent
\begin{figure*}[h]
    \begin{center}
    \centering
    \includegraphics[width=1\textwidth]{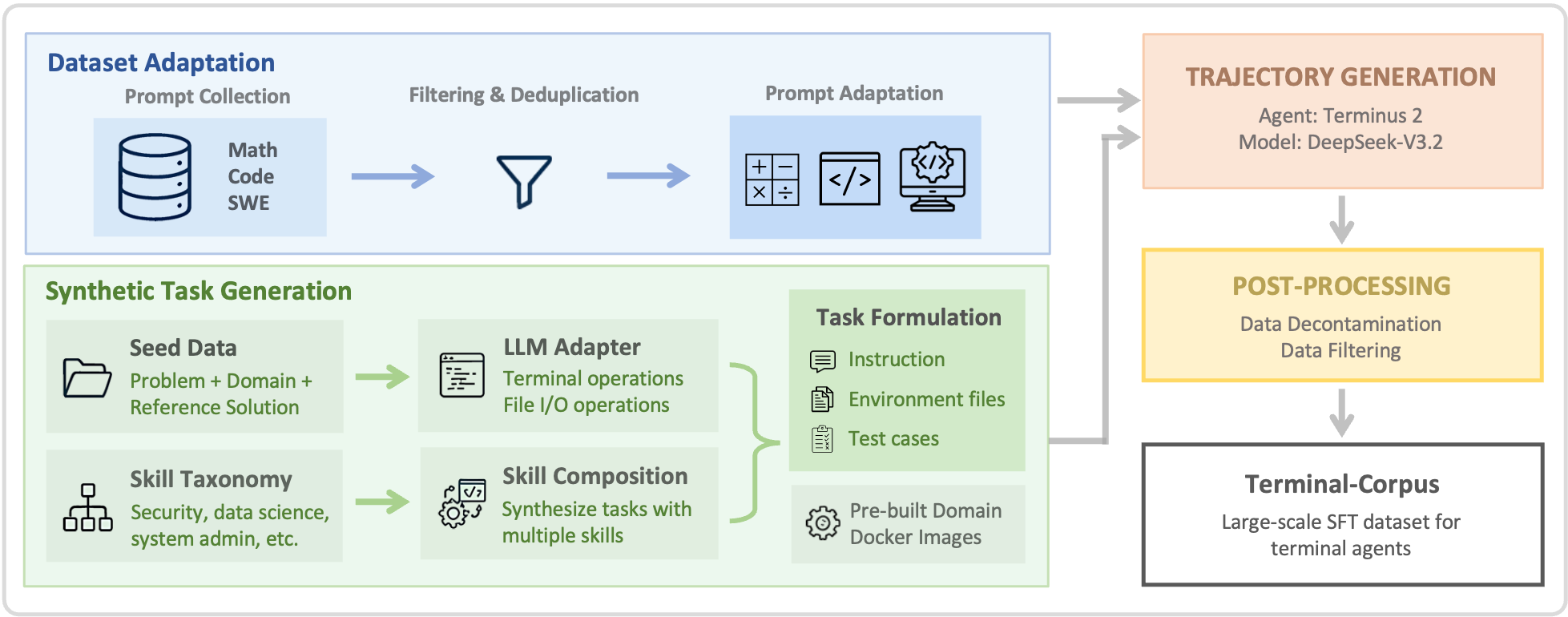}
    \caption{\textbf{Overview of \ourmethod.} Our framework combines Dataset Adaptation, which transforms existing benchmarks into terminal prompts, with Synthetic Task Generation, which uses seed data and a Skill Taxonomy to construct targeted scenarios. The tasks from both streams are utilized during Trajectory Generation phase, where agents interact with Dockerized environments to produce solution traces, followed by Post-Processing (decontamination and filtering) to yield the final SFT dataset.
    }
    \label{fig:overview}
    \end{center}
\end{figure*}

\section{Introduction}

As large language models (LLMs) advance toward practical software engineering applications, terminal interaction has emerged as a critical capability. Tools like Claude Code \citep{anthropic2025claudecode} and Codex CLI \citep{openai2025codex} demonstrate the potential of command-line proficiency, with frontier models showing promising results on benchmarks such as Terminal-Bench \citep{merrill2026terminal}. However, the training data mixtures behind these systems remain largely undisclosed, leaving fundamental questions about effective data design unanswered. This lack of transparency forces researchers into a costly trial-and-error process, primarily due to two significant bottlenecks in agentic data generation: (1) the scarcity of foundational resources, including diverse task prompts, requisite dependency files, and pre-configured environments; and (2) the logistical complexity of trajectory collection, as real-world human interactions are difficult to capture, while synthetic generation via LLM agents is prohibitively expensive due to the need for fresh environment instantiation and multi-turn interaction for every task.
\begin{table}[t!]
    \centering
    \caption{\textbf{Model performance comparison on Terminal-Bench 2.0.}}
    \label{tab:model_comparison_split}
    \resizebox{\textwidth}{!}{%
        \begin{tabular}{lcccccc}
            \toprule
            \multicolumn{7}{c}{\textbf{Closed Source Models}} \\
            \midrule
            \textbf{Model} & GPT-5-Nano & GPT-5-Mini & Grok Code Fast 1 & Grok 4 & Gemini 2.5 Flash & Qwen3-Max-Thinking \\
            \textbf{Accuracy} & 7.90 & 24.0 & 14.2 & 23.1 & 16.9 & 22.5 \\
            \midrule
            \multicolumn{7}{c}{\textbf{Open Source Models}} \\
            \midrule
            \textbf{Model} & Qwen3-32B & Qwen3-Coder & GPT-OSS (high) 20B & GPT-OSS (high) 120B & Nemotron-T-14B & Nemotron-T-32B \\
            \textbf{Size} & 32B & 480B & 20B & 120B & 14B & 32B \\
            \textbf{Accuracy} & 3.37 & 23.9 & 3.10 & 18.7 & \textbf{20.2} & \textbf{27.4} \\
            \bottomrule
        \end{tabular}%
    }
\end{table}
Current approaches to improving terminal capabilities fall into two main categories: improving agentic scaffolds \citep{singhal2025droid, antigma2025ante, jetbrains2025junie, ii2025iiagent, nichols2025terminalbenchwarp, mux2025terminalbenchmarking, letta2025terminalbench} or improving the underlying model for terminal use \citep{openai2025gpt52, google2025gemini3, anthropic2025claudeopus45, liu2025deepseek, moonshotai2025kimik2thinking, minimaxM2_2025}. One prominent approach for post-training models involves using adapters that wrap existing datasets in command-line interfaces \citep{mlfoundations2025staqcterminus2, mlfoundations2025terminus2codecontests, dcagent2025bashtextbook}. While the Terminal-Bench authors provide a repository of such adapters \citep{harbor2025adapters} primarily for benchmark adaptation, these adapters can also be repurposed as a starting point for scaling training data. However, adapters inherit structural assumptions from source datasets never designed for sequential environment interaction, potentially limiting their effectiveness. Recent work has explored multi-agent frameworks \citep{danau5tin2025tbenchpipeline, peng2025litecoderterminal} for more principled data generation, but these introduce computational complexity that scales poorly for large-scale training. The field thus lacks a practical framework that balances generation efficiency with the specific requirements of training effective terminal agents.

We address this gap through a dual-strategy approach that combines \emph{dataset adaptation} with \emph{synthetic task generation}. Dataset adaptation provides broad coverage by transforming existing math, code, and software engineering datasets into Terminal-Bench format, efficiently scaling data volume while leveraging high-quality problem sources. Synthetic task generation offers finer-grained control: it enables targeted development of terminal-specific skills with control over task characteristics, such as difficulty, domain coverage, and primitive skill composition. These strategies are complementary and address distinct bottlenecks: adapters make the most of existing datasets to build foundational terminal capabilities at scale, while synthetic task generation provides the flexibility to target specific capability gaps. Building on this coarse-to-fine data generation pipeline, we develop \textbf{\ourmodel}, a family of models fine-tuned from Qwen3 models \citep{yang2025qwen3}.

Specifically, we make the following contributions:

\begin{enumerate}
  \item We introduce \textbf{\ourmethod}, a scalable synthetic task generation pipeline that enables rapid exploration of task design and targeted generation for specific skills with different difficulty levels.
  \item We conduct a \textbf{systematic study of data engineering strategies}, examining filtering strategies for dataset adapters and synthetic tasks, curriculum learning for data mixing, long context training, and scaling trends.
\item We conduct extensive experiments to validate the effectiveness of our curated dataset. As demonstrated in Table~\ref{tab:model_comparison_split}, on Terminal-Bench 2.0, our models achieve substantial improvements over the initial Qwen3 baselines, reaching competitive performance with significantly larger models while requiring only modest computational resources for training and inference. For example, our \ourmodel-32B outperforms Qwen3-Coder-480B~\citep{yang2025qwen3} on Terminal-Bench 2.0~(27.4$\pm$ 2.4 vs. 23.9$\pm$ 2.8). We release \textbf{\ourmodel models and \ourdata dataset} at \url{https://huggingface.co/collections/nvidia/nemotron-terminal}.
\end{enumerate}

\section{Related Work}

\paragraph{Agent Design.} As seen through Claude Code \citep{anthropic2025claudecode} and Codex CLI \citep{openai2025codex}, sophisticated agent scaffolding can significantly improve performance. Many leading terminal agents achieve frontier performance through innovations in scaffolding \citep{singhal2025droid, antigma2025ante, jetbrains2025junie, ii2025iiagent, nichols2025terminalbenchwarp, mux2025terminalbenchmarking, letta2025terminalbench}. However, effective agent scaffolds are often model-specific and require extensive engineering. As base models improve, the marginal benefit of complex scaffolding will likely decrease. Rather than exploring variants in agentic design, we focus on scaling underlying model capabilities through targeted supervised fine-tuning.

\paragraph{Dataset Adapters.} Several datasets on Hugging Face \citep{mlfoundations2025staqcterminus2, mlfoundations2025terminus2codecontests, dcagent2025bashtextbook} collect agent execution traces by rolling out prompts from existing datasets through terminal environments. This approach can efficiently scale data collection by reusing existing prompts from different domains, including competitive coding and math. Despite the abundance of these datasets, no formal analysis has studied which characteristics of dataset adapters can affect downstream training effectiveness. In this work, we explore the strengths and weaknesses of this approach through a systematic study using custom datasets and adapters.

\paragraph{Synthetic Task Generation.} Many studies have investigated how to effectively generate synthetic data for fine-tuning LLMs. Evol-Instruct \citep{xu2023wizardlm} pioneered automated instruction data scaling through iterative in-depth and in-breadth evolution. Code Evol-Instruct \citep{luo2023wizardcoder} successfully adapted this strategy to automatically increase the complexity of code instruction data for effective fine-tuning. Since then, AgentInstruct \citep{mitra2024agentinstruct} and LAB \citep{sudalairaj2024lab} have demonstrated how to generate large-scale datasets from existing seed data through suggester-editor agent pairs and taxonomy-driven generation. Other works like MAGPIE \citep{xu2024magpie} have explored extracting instruction data from aligned LLMs without seed data through unique prompting tactics. 

Recent work has explored bringing these ideas to scale terminal capabilities in LLMs, employing multi-agent systems to brainstorm ideas, generate tasks, design Docker environments, and validate generated tasks and environments \citep{danau5tin2025tbenchpipeline, peng2025litecoderterminal}. Since multi-agent systems can be time-consuming and costly, we design a simplified system that eliminates unnecessary coordination stages and optimizes environment validation to enable effective scaling. Through our systematic study and ablations, we provide actionable insights for scaling terminal-capable models.
\section{Background}

\subsection{Terminal-Bench}

Terminal-Bench \citep{merrill2026terminal, tbench_2025} has emerged as the standard benchmark for evaluating agents in terminal environments. The benchmark comprises 89 hand-crafted, human-verified tasks that span diverse domains including scientific computing, software engineering, machine learning, security, system administration, and data science. Unlike traditional code generation benchmarks that evaluate isolated functions, Terminal-Bench tasks require agents to complete end-to-end workflows, such as compiling code, training models, configuring systems, and debugging environments.

As seen in Figure~\ref{fig:task-structure}, each task in Terminal-Bench includes four components: (1) a natural language instruction describing the objective, (2) a containerized Docker \citep{merkel2014docker} environment providing the execution context, (3) a verification test suite that programmatically checks task completion, and (4) an oracle solution demonstrating a valid approach. Throughout this work, we use Terminal-Bench 2.0 as our primary evaluation benchmark and leverage Terminus 2, the model-agnostic reference agent released alongside the benchmark, for consistent evaluation across model checkpoints.

\begin{figure}[htbp]
    \small
    \centering
    \begin{minipage}[t]{0.48\textwidth}
        \centering
        \begin{codebox}
task_directory/
  instruction.md
  task.toml
  environment/
    Dockerfile
    ...
  solution/
    solve.sh
    ...
  tests/
    test.sh
    ...
        \end{codebox}
        \caption{\textbf{Terminal-Bench task directory structure.} Each task consists of an instruction prompt, task metadata, environment files, Dockerfile, reference solution, and test cases.}
        \label{fig:task-structure}
    \end{minipage}
    \hfill 
    \begin{minipage}[t]{0.48\textwidth}
        \centering
        \begin{codebox}[language=json]
{
  "analysis": "...",
  "plan": "...",
  "commands": [
    {
      "keystrokes": "ls -la\n",
      "duration": 0.1
    },
    {
      "keystrokes": "cd project\n",
      "duration": 0.1
    }
  ],
  "task_complete": true
}
        \end{codebox}
        \caption{\textbf{Terminus 2 agent response format.} The Terminus 2 agent scaffold prompts the model to output responses in a JSON format, which includes: \textit{analysis}, \textit{plan}, \textit{commands}, and \textit{task\_complete}.}
        \label{fig:terminus-format}
    \end{minipage}
\end{figure}


\subsection{Terminus 2 Agent Framework}

Unlike traditional coding agents that provide multiple specialized tools, Terminus 2 \citep{tbench_2025} only provides an interactive tmux session running inside a sandboxed Docker \citep{merkel2014docker} container. Through sending model-determined keystrokes to the tmux session, the agent has the flexibility to approach tasks using any available command-line tools.

At each step, the agent receives the current terminal output, and the model is prompted to respond with a structured JSON format (Figure~\ref{fig:terminus-format}) that determines the next action sent to the environment.


\section{Synthetic Data Generation}

Training autonomous agents for terminal environments requires a systematic approach to data curation that balances breadth, depth, and scalability. We introduce a principled two-stage data generation framework: \emph{dataset adaptation} for establishing broad foundational coverage, followed by \emph{synthetic task generation} for targeted skill refinement. This coarse-to-fine strategy decouples data volume scaling from task design iteration: adapters efficiently leverage existing problem repositories to build general competencies, while synthetic generation enables precise control over skill composition, difficulty progression, and domain-specific requirements. Together, these complementary strategies yield a diverse, high-quality dataset for supervised fine-tuning (SFT) that systematically covers the operational and computational skills required for terminal-based problem solving.

\subsection{Dataset Adapters}

\subsubsection{Prompt Datasets}

We selectively identify targeted high quality SFT prompt datasets that span the math, code, and software engineering (SWE) domains, since they are foundational to several of the topics covered in terminal use. 

\paragraph{Math Prompts.} We use the Stage-2 prompt set from Nemotron-Cascade's math reasoning SFT data \citep{wang2025nemotron}, which consists of 163K unique prompts drawn from OpenMathReasoning \citep{moshkov2025aimo}. To obtain this high-quality prompt set, Nemotron-Cascade filters out easy questions from the original datasets by excluding prompts whose DeepSeek-R1 \citep{guo2025deepseek} response length is shorter than 2K tokens.

\paragraph{Code Prompts.} We use the Stage-2 prompt set from Nemotron-Cascade's code reasoning SFT data, which consists of 79K prompts from OpenCodeReasoning \citep{ahmad2025opencodereasoning} covering challenging coding problems. We further filter and deduplicate this set to obtain a 35K prompt subset.

\paragraph{SWE Prompts.} For software engineering tasks, we draw from Nemotron-Cascade's SWE code repair SFT data, which consists of 127K instances from SWE-Bench-Train \citep{jimenez2023swe}, SWE-reBench \citep{badertdinov2025swe}, SWE-Smith \citep{yang2025swe}, and SWE-Fixer-Train \citep{xie2025swe}. Each prompt includes a problem statement and the contents of one or more buggy code files. We further filter and deduplicate this set, resulting in 32K unique prompts.

\subsubsection{Adapter Format}

Dataset adaptation is a straightforward process that converts existing prompt datasets to Terminal-Bench format without requiring an LLM in the loop. Using the Terminus 2 system prompt template (Appendix~\ref{appendix:trajectory-generation-templates}), we map each entry to the \textit{\{instruction\}} placeholder, appending a unique instruction suffix based on the dataset type. The specific suffixes used for math, code, and SWE prompts are detailed in Appendix~\ref{appendix:trajectory-generation-templates}. For each code file identified in a SWE prompt, we instantiate a corresponding file within the environment. As the Nemotron-Cascade datasets provide only prompts, these tasks consist of an instruction and environment, without associated test cases.

\subsection{Synthetic Task Generation}
\label{sec:synthetic-task-generation}
While dataset adapters provide a foundational breadth of reasoning and code, they are inherently limited by the formats of their source repositories. To bridge the gap between general problem-solving and the specific rigors of terminal-based agency, we introduce \textbf{\ourmethod}, a synthetic pipeline that generates executable tasks with precise control over skill complexity and environment constraints. By generating tasks from both structured seeds and a taxonomy of primitive skills, we ensure the training data directly reflects the operational nuances and multi-step tool interactions required for terminal agents.

We present two complementary approaches for generating synthetic terminal operation tasks: \textbf{seed-based generation} and \textbf{skill-based generation}. Both methods leverage LLMs to produce diverse, executable terminal tasks while addressing distinct requirements of scalability, diversity, and domain coverage.

\subsubsection{Generation from Seed Data}

Seed-based generation complements dataset adaptation by using existing problems as inspiration, rather than as fixed templates. Instead of wrapping original prompts in a terminal scaffold, we prompt an LLM to synthesize new terminal tasks from seed problems. This approach is particularly effective for leveraging high-quality problem specifications from adjacent domains, such as scientific computing challenges, algorithmic problem sets, or domain-specific coding exercises, where well-defined problems exist but lack the terminal-oriented task structure required for agent training.

\paragraph{Seed Data Structure.}
Each seed entry is a structured record containing: (1)~a \emph{problem description} specifying the computational challenge, (2)~an optional \emph{domain label} indicating the scientific or technical area (e.g., biology, physics, optimization), and (3)~an optional \emph{reference solution} providing a correct implementation. The reference solution, when available, serves as ground truth for generating test expectations but is never exposed to the agent.

\paragraph{Task Adaptation.}
The LLM acts as a \emph{task adapter} that transforms each seed problem into a self-contained terminal task. This transformation involves several key operations. First, the abstract problem statement is augmented with concrete software engineering requirements: the agent must install necessary packages, read input from specified file paths, implement the solution, and write results to designated output locations. Second, the adapter generates realistic input data files that instantiate the problem with specific test cases, including edge cases and boundary conditions. Third, comprehensive pytest-based test cases are synthesized to verify correctness, which check output file existence, format compliance, numerical accuracy (with appropriate tolerances for floating-point results), and edge case handling. When a reference solution is provided in the seed data, it is included in the generation context, and the prompt instructs the LLM to use it as ground truth when designing the test cases.

\paragraph{Conversion Guidelines.}
The conversion prompt encodes several principles to ensure task quality: complex problems are decomposed into verifiable units when necessary; practical constraints such as input sizes and precision requirements are added to ground the problem in realistic scenarios; and output formats are designed to enable unambiguous programmatic verification. This systematic adaptation enables the pipeline to convert diverse problem sources into a uniform task format suitable for agent evaluation.

\subsubsection{Generation from Primitive Skills}

Skill-based generation takes a fundamentally different approach: rather than adapting existing problems, it synthesizes novel tasks from a structured taxonomy of primitive terminal operation skills. We curate a list of primitive terminal operation skills and employ LLMs to expand and recombine these primitives into creative task specifications.

\paragraph{Domain-Specific Generation.}
The task generation process is inherently domain-specific. We define 9 task domains: data processing, data querying, data science, debugging, dependency management, file operations, scientific computing, security, and software engineering. Each domain is associated with a dedicated generation prompt that guides the LLM to produce tasks aligned with the domain's focus areas. For instance, the data science prompt directs the model toward tasks involving statistical analysis and data transformation, whereas the security prompt emphasizes cryptographic operations and access control verification. This domain-aware prompting ensures that generated tasks exhibit coherent thematic focus while exercising skills appropriate to the target category.

\paragraph{Skill Taxonomy.}
In each domain, primitive skills are collected and span multiple dimensions of terminal-based problem solving: (1)~\emph{algorithmic} skills such as graph traversal, constraint satisfaction, and backtracking search; (2)~\emph{systems} skills including file I/O, process management, and network configuration; (3)~\emph{data processing} skills such as parsing, serialization, and transformation pipelines; (4)~\emph{mathematical} skills including numerical integration and statistical modeling; (5)~\emph{testing} skills such as validation, verification, and benchmarking; and (6)~\emph{web/security} skills including HTTP handling, authentication, and vulnerability analysis. Skills for each domain are summarized in Appendix.

\paragraph{Compositional Task Synthesis.}
The LLM is instructed to combine multiple primitives (typically 3--5 skills per task) in non-trivial ways, producing tasks that require integrated problem-solving rather than isolated skill application. Crucially, the generation prompt emphasizes \emph{novelty}: the model is guided to invent new scenarios, thereby maximizing the diversity and coverage of the resulting tasks.

\subsubsection{Task Format and Execution Environment}

Both generation methods produce tasks in a standardized format comprising: (1)~a natural language task prompt specifying objectives and constraints, (2)~pytest-based test cases with configurable weights for partial credit, (3)~supplementary input files providing necessary data, and (4)~a domain-specific Docker environment for consistent execution. The files are structured in the same way as Terminal-Bench, as shown in Figure~\ref{fig:task-structure}.
Note we do not generate oracle solutions, as producing ground-truth code is prohibitively difficult without human verification; instead, we generate tasks are easy to verify yet difficult to solve and use the synthesized test cases to evaluate the correctness of agent solutions.

\paragraph{Solution Isolation.}
A key design principle enforced across all generation prompts is the separation between problem specification and solution information. All prompts explicitly instruct the LLM to avoid solution leakage: the task prompt visible to the agent must not reveal the algorithm, implementation approach, or any code that solves the problem. When reference solutions are available (e.g., in seed data), they are used exclusively for deriving ground-truth test expectations. This ensures that generated tasks require problem-solving rather than simple solution retrieval.

\paragraph{Pre-Built Docker Images.}
A critical design decision that enables large-scale task generation is the use of \textbf{pre-built, domain-specific Docker images}. Rather than generating a unique Dockerfile per task similar to previous works~\citep{danau5tin2025tbenchpipeline, peng2025litecoderterminal}, we maintain a fixed set of domain specific docker images, each pre-installing the packages and dependencies commonly required within that domain (e.g., pandas and scikit-learn for data science; cryptography libraries for security).

This approach provides three scalability advantages. First, it \emph{eliminates Dockerfile validation overhead}; by avoiding the costly multi-turn repair often needed for per-task environment generation, pre-built images enable efficient single-pass task creation. Second, it \emph{reduces resource footprint}, utilizing just 9 shared base images instead of building and caching thousands of unique containers. Third, it \emph{decouples environment and task generation}, allowing the pipeline to produce diverse scenarios within stable environments while retaining the flexibility for agents to install runtime dependencies.


\subsection{Teacher Model}

We select DeepSeek-V3.2 \citep{liu2025deepseek} as our teacher model for generating synthetic tasks and trajectories, motivated by its strong performance on Terminal-Bench 2.0 (Table~\ref{tab:terminalbench}). To further validate its suitability for producing dataset adapter trajectories, we evaluate DeepSeek-V3.2 on a few standard benchmarks adapted to Terminal-Bench format using the Terminus 2 agent framework (Table~\ref{tab:proxy-benchmark-results}), including AIME 2024 \citep{aime2024-benchmark}, AIME 2025 \citep{aime2025-benchmark}, LiveCodeBench v6 \citep{jain2024livecodebench}, and SWE-bench Verified \citep{jimenez2023swe, openai2024swebenchverified}. 

\begin{table}[t]
  \caption{\textbf{DeepSeek-V3.2 performance on adapted math, coding, and SWE benchmarks.} Using the Terminus 2 agent, we evaluate DeepSeek-V3.2 on AIME, LiveCodeBench, and SWE-bench Verified adapted to Terminal-Bench format, and we find that it performs reasonably well even under this terminal-based setting.}
  \label{tab:proxy-benchmark-results}
  \begin{center}
    \begin{small}
      \begin{sc}
        \begin{tabular}{lcc}
          \toprule
          Benchmark (pass@1) & DeepSeek-V3.2 \\
          \midrule
          AIME 2024, AIME 2025 & 93.33 \\
          LiveCodeBench v6 & 67.20 \\
          SWE-bench Verified & 52.40 \\
          \bottomrule
        \end{tabular}
      \end{sc}
    \end{small}
  \end{center}
  \vskip -0.1in
\end{table}

\subsection{Data Filtering}
\label{sec:data_filtering}

We first decontaminate our SFT dataset by removing any prompt that has a 14-gram overlap with Terminal-Bench 2.0 test samples. Then, we apply various quality filters, including removing identity leaks and discarding responses that contain Chinese characters. 

Beyond quality filtering, we also experiment with removing incomplete trajectories generated by the teacher model to discourage the fine-tuned model from becoming overly verbose. When tests are available, we further experiment with only keeping trajectories that pass the tests.

\section{Experiments}

\subsection{Experimental Setup}

\textbf{Base Models.} We conduct experiments using pretrained models from the Qwen3 family \citep{yang2025qwen3}. We use Qwen3-8B as our primary model for ablation studies, and we additionally experiment with Qwen3-14B and Qwen3-32B to verify that our findings scale with model size.

\begin{wraptable}{r}{6.8cm} 
\centering
\vspace{-7mm}
\caption{\textbf{Terminal-Bench 2.0 (TB2.0) results with Terminus 2 agent.}}
\label{tab:terminalbench}
\resizebox{\linewidth}{!}{%
    \begin{tabular}{l|c|c} 
    \hline 
    \textbf{Model} & \textbf{Size} & \textbf{TB2.0} \\ 
    \hline
    \multicolumn{3}{c}{\textit{Closed Source Models}} \\ 
    \hline
    GPT-5-Nano & -- & 7.90 $\pm$ 1.9 \\
    GPT-5-Mini & -- & 24.0 $\pm$ 2.5 \\
    GPT-5 & -- & 35.2 $\pm$ 3.1 \\
    GPT-5.1 & -- & 47.6 $\pm$ 2.8 \\
    GPT-5.2 & -- & 54.0 $\pm$ 2.9 \\
    Grok Code Fast 1 & -- & 14.2 $\pm$ 2.5 \\
    Grok 4 & -- & 23.1 $\pm$ 2.9 \\
    GLM 4.6 & -- & 24.5 $\pm$ 2.4 \\
    Gemini 2.5 Flash & -- & 16.9 $\pm$ 2.4 \\
    Gemini 2.5 Pro & -- & 32.6 $\pm$ 3.0 \\
    Gemini 3 Flash & -- & 51.7 $\pm$ 3.1 \\
    Gemini 3 Pro & -- & 56.9 $\pm$ 2.5 \\
    Claude Haiku 4.5 & -- & 28.3 $\pm$ 2.9 \\
    Claude Sonnet 4.5 & -- & 42.8 $\pm$ 2.8 \\
    Claude Opus 4.5 & -- & 57.8 $\pm$ 2.5 \\
    \hline
    \multicolumn{3}{c}{\textit{Open Source Models}} \\
    \hline
    Qwen3-8B & 8B & 2.47 $\pm$ 0.5 \\
    Qwen3-14B & 14B & 4.04 $\pm$ 1.3 \\
    Qwen3-32B & 32B & 3.37 $\pm$ 1.6 \\
    Qwen3-Coder & 480B & 23.9 $\pm$ 2.8 \\
    GPT-OSS (high) & 20B & 3.10 $\pm$ 1.5 \\
    GPT-OSS (high) & 120B & 18.7 $\pm$ 2.7 \\
    MiniMax M2 & 230B & 30.0 $\pm$ 2.7 \\
    MiniMax M2.1 & 230B & 29.2 $\pm$ 2.9 \\
    Kimi K2 Thinking & 1T & 35.7 $\pm$ 2.8 \\
    DeepSeek-V3.2 & 685B & 38.2 $\pm$ 2.9 \\
    \hline
    \multicolumn{3}{c}{\textit{Ours}} \\
    \hline
    \ourmodel-8B & 8B & 13.0 $\pm$ 2.2 \\
    \ourmodel-14B & 14B & 20.2 $\pm$ 2.7 \\
    \textbf{\ourmodel-32B} & 32B & \textbf{27.4 $\pm$ 2.4} \\
    \hline
    \end{tabular}%
}
\vspace{-16mm}
\end{wraptable}

\textbf{Training Details.} Unless specified otherwise, we use the following training hyperparameters: learning rate of 2e-5, weight decay of 1e-4, 2 epochs, maximum sequence length of 32,768 tokens, global batch size of 128, and micro-batch size of 1 per GPU, with AdamW optimizer ($\beta$ = 0.9, 0.95), cosine learning rate scheduler with 10\% warmup, and gradient clipping at 1.0. The 8B and 14B models are trained on 4 nodes with 8 GPUs per node (32 total GPUs) using sequence parallelism of 2. The 32B model is trained on 16 nodes (128 total GPUs). All experiments use CPU offloading.


\textbf{Infrastructure.} We utilize Harbor \citep{Shaw_Harbor_Framework_2025}, the infrastructure framework from Terminal-Bench 2.0 \citep{tbench_2025}, to orchestrate large-scale trajectory generation in containerized environments. We extend Harbor to support Singularity \citep{kurtzer2017singularity}, enabling deployment on HPC clusters; while this introduces rare failures due to fakeroot overlay limitations, these are acceptable for synthetic data generation. For evaluation, we rely on Daytona \citep{daytona_cloud_2025} to manage reliable, parallel execution in isolated cloud sandboxes.

For SFT experiments, we use veRL \citep{sheng2024hybridflow}, an open-source framework designed for efficient LLM training.

\subsection{Main Results}
We evaluate \ourmethod by benchmarking \ourmodel on Terminal-Bench 2.0 (TB2.0). As shown in Table~\ref{tab:terminalbench}, our models demonstrate substantial gains: \ourmodel-8B achieves $13.0 \pm 2.2$, a five-fold increase over Qwen3-8B ($2.47 \pm 0.5$). Remarkably, despite their modest scale, they rival significantly larger systems; \ourmodel-14B ($20.2 \pm 2.7$) outperforms the 120B GPT-OSS ($18.7 \pm 2.7$) and Gemini 2.5 Flash ($16.9 \pm 2.4$), while \ourmodel-32B ($27.4 \pm 2.4$) outperforms the 480B Qwen3-Coder ($23.9 \pm 2.8$). This validates that high-quality trajectory data effectively bridges the gap between efficient models and massive frontier counterparts.

Further analysis by task category (Table~\ref{tab:qwen3_base_category}) confirms that our synthetic data unlocks critical capabilities where base models of all scales failed completely. While the Qwen3-14B and 32B models both scored 0.0 in Data Querying and Model Training, \ourmodel-32B surged to 60.0 and 50.0 respectively, representing a profound leap in functional utility. Similar transformations occur in Security (2.5 to 27.5), Data Processing (5.0 to 50.0), and Software Engineering (5.0 to 31.7) for the 32B variant, indicating that larger parameter count alone is insufficient for strong terminal capability. Furthermore, consistent gains in System Administration (6.7 to 31.1) and Debugging (0.0 to 33.3) demonstrate that our approach successfully instills domain-specific skills—such as complex file manipulation and command-line troubleshooting, which are effectively absent in the original models.

\begin{table}[h]
\centering
\caption{\textbf{Performance of models by Terminal-Bench category.} TB2.0 scores for Qwen3 and \ourmodel models, broken down by category, showing where \ourmodel most improves over the Qwen3 baselines.}
\label{tab:qwen3_base_category}
\resizebox{0.7\linewidth}{!}{%
    \begin{tabular}{l:ccc:ccc}
    \hline
    \textbf{Category} & \multicolumn{3}{c:}{\textbf{Qwen3}} & \multicolumn{3}{c}{\textbf{\ourmodel}} \\
    \cline{2-7}
     & \textbf{8B} & \textbf{14B} & \textbf{32B} & \textbf{8B} & \textbf{14B} & \textbf{32B} \\
    \hline
    \multicolumn{7}{l}{\textbf{Software \& System}} \\
    \hspace{3mm} Software Engineering (24) & 1.70 & 6.70 & 5.00 & 9.20 & 18.3 & 31.7 \\
    \hspace{3mm} System Administration (9) & 13.3 & 6.70 & 6.70 & 22.2 & 28.9 & 31.1 \\
    \hspace{3mm} Debugging (3) & 0.00 & 0.00 & 0.00 & 20.0 & 40.0 & 33.3 \\
    \hspace{3mm} Security (8) & 0.00 & 0.00 & 2.50 & 12.5 & 17.5 & 27.5 \\
    \hspace{3mm} File Operations (4) & 0.00 & 0.00 & 0.00 & 0.00 & 10.0 & 5.00 \\
    \hline
    \multicolumn{7}{l}{\textbf{Data \& Science}} \\
    \hspace{3mm} Data Science (8) & 2.50 & 7.50 & 0.00 & 7.50 & 17.5 & 27.5 \\
    \hspace{3mm} Data Processing (4) & 0.00 & 5.00 & 5.00 & 35.0 & 40.0 & 50.0 \\
    \hspace{3mm} Data Querying (1) & 0.00 & 0.00 & 0.00 & 20.0 & 40.0 & 60.0 \\
    \hspace{3mm} Scientific Computing (7) & 0.00 & 0.00 & 2.90 & 0.00 & 2.90 & 0.00 \\
    \hspace{3mm} Mathematics (4) & 0.00 & 0.00 & 0.00 & 0.00 & 0.00 & 0.00 \\
    \hline
    \multicolumn{7}{l}{\textbf{Machine Learning}} \\
    \hspace{3mm} Machine Learning (3) & 0.00 & 0.00 & 0.00 & 6.70 & 13.3 & 13.3 \\
    \hspace{3mm} Model Training (4) & 0.00 & 0.00 & 0.00 & 5.00 & 20.0 & 50.0 \\
    \hline
    \multicolumn{7}{l}{\textbf{Other}} \\
    \hspace{3mm} Personal Assistant (1) & 0.00 & 0.00 & 0.00 & 80.0 & 80.0 & 100 \\
    \hspace{3mm} Games (1) & 0.00 & 0.00 & 0.00 & 0.00 & 0.00 & 0.00 \\
    \hspace{3mm} Video Processing (1) & 0.00 & 0.00 & 0.00 & 0.00 & 0.00 & 0.00 \\
    \hspace{3mm} Unknown (7) & 5.70 & 8.60 & 8.60 & 34.3 & 34.3 & 34.3 \\
    \hline
    \textbf{Overall} & 2.50 & 4.00 & 3.40 & 13.0 & 20.2 & 27.4 \\
    \hline
    \end{tabular}%
}
\end{table}


\begin{table}[h!]
    \small
    \centering
    \begin{minipage}[t]{0.48\textwidth}
        \centering
        \caption{\textbf{Qwen3-8B SFT results across training data sources.} We show total samples and TB2.0 results from training on various subsets. For both dataset adapters and synthetic tasks, we achieve the strongest results from combining all data sources together.}
        \label{tab:data_source_ablation}
        \vspace{5pt}
        \begin{tabular}[t]{lcc}
            \hline
            \textbf{Data Split} & \textbf{\# Samples} & \textbf{TB2.0} \\
            \hline
            \multicolumn{3}{l}{\textbf{Dataset Adapters}} \\
            \hspace{3mm} Math & 162,692 & 5.39 $\pm$ 1.65 \\
            \hspace{3mm} Code & 31,960 & 6.29 $\pm$ 1.65 \\
            \hspace{3mm} SWE  & 31,661 & 7.02 $\pm$ 2.13 \\
            \hspace{3mm} All  & 226,313 & \textbf{9.66 $\pm$ 2.11} \\
            \hline
            \multicolumn{3}{l}{\textbf{Synthetic Tasks}} \\
            \hspace{3mm} Seed-based  & 124,366 & 6.18 $\pm$ 1.91 \\
            \hspace{3mm} Skill-based & 139,841 & 12.4 $\pm$ 2.38 \\
            \hspace{3mm} All         & 264,207 & \textbf{12.4 $\pm$ 2.29} \\
            \hline
        \end{tabular}
    \end{minipage}
    \hfill 
    \begin{minipage}[t]{0.48\textwidth}
        \centering
        \caption{\textbf{Qwen3-8B ablations on dataset adapter filtering strategies.} Because the dataset adapter subset does not include test cases, we compare no filtering to complete-only filtering and observe no significant performance difference.}
        \label{tab:dataset_adapter_filtering}
        \vspace{-5pt}
        \begin{tabular}[t]{lrr}
            \toprule
            \textbf{Filter} & \textbf{\# Samples} & \textbf{TB2.0} \\
            \midrule
            \multicolumn{3}{l}{\textbf{Math}} \\
            \quad Complete-only & 147,718 & \textbf{7.19 $\pm$ 1.87} \\
            \quad No filter     & 162,692 & 5.39 $\pm$ 1.65 \\
            \midrule
            \multicolumn{3}{l}{\textbf{Code}} \\
            \quad Complete-only & 20,169  & 6.07 $\pm$ 1.73 \\
            \quad No filter     & 31,960  & \textbf{6.29 $\pm$ 1.65} \\
            \midrule
            \multicolumn{3}{l}{\textbf{SWE}} \\
            \quad Complete-only & 29,053  & 5.39 $\pm$ 1.68 \\
            \quad No filter     & 31,661  & \textbf{7.02 $\pm$ 2.13} \\
            \midrule
            \multicolumn{3}{l}{\textbf{All}} \\
            \quad Complete-only & 196,940 & 8.09 $\pm$ 1.84 \\
            \quad No filter     & 226,313 & \textbf{9.66 $\pm$ 2.11} \\
            \bottomrule
        \end{tabular}
    \end{minipage}
\end{table}


\subsection{Ablation on Dataset Components}
As demonstrated in Table~\ref{tab:data_source_ablation}, each data source provides complementary value. For dataset adapters, while individual splits like Math (5.39\%) and Code (6.29\%) underperform compared to SWE (7.02\%), combining them leads to a significant performance jump to 9.66\%, confirming that merging distinct domains yields further improvement than any single source alone. A similar pattern of robustness appears in synthetic tasks, where the skill-based data drives the primary gains (12.4\%); although adding seed-based data does not increase the mean score, it successfully reduces variance and makes the model more robust.


\subsection{Filtering Strategies}
\paragraph{Trajectory Filtering} We investigate a few trajectory filtering strategies discussed in Section~\ref{sec:data_filtering}. For dataset adapters (Table~\ref{tab:dataset_adapter_filtering}), while we observe no significant difference on each subset, the no-filter setting yields the highest performance on the full set (9.66\%) and is thus adopted. For synthetic tasks (Table~\ref{tab:synthetic_task_filtering}), the impact is even more substantial: no filtering (12.4\%) significantly surpasses both complete-only (6.74\%) and success-only (5.06\%) strategies. This performance gap suggests that strict filtering is detrimental as it discards over half the available training data. Moreover, retaining unsuccessful trajectories appears to provide valuable supervision, exposing the model to realistic error states and recovery patterns that enhance overall robustness.

\begin{table}[h]
    \small
    \centering
    \begin{minipage}[t]{0.45\textwidth}
        \centering
        \caption{\textbf{Qwen3-8B ablations on synthetic task filtering.} On the synthetic task subset, we experiment with no filtering, complete-only filtering, and success-only filtering. No filtering yields significantly better performance.}
        \label{tab:synthetic_task_filtering}
        \vspace{7pt} 
        \begin{tabular}[t]{lrr}
            \hline
            \textbf{Filter} & \textbf{\# Samples} & \textbf{TB2.0} \\
            \hline
            Complete-only & 104,603 & 6.74 $\pm$ 2.20 \\
            Success-only  & 83,448  & 5.06 $\pm$ 2.11 \\
            No filter     & 264,207 & \textbf{12.4 $\pm$ 2.29} \\
            \hline
        \end{tabular}
    \end{minipage}
    \hfill 
    \begin{minipage}[t]{0.52\textwidth}
        \centering
        \caption{\textbf{Impact of SFT sequence length and YaRN2 scaling on Qwen3-8B performance.} We find that training and evaluating with default Qwen3 context settings yields the strongest performance.}
        \label{tab:long_context}
        \vspace{-3pt}
        \resizebox{\textwidth}{!}{%
        \begin{tabular}[t]{lcccc}
            \toprule
            \textbf{SFT Max} & \textbf{Eval Max} & \textbf{SFT} & \textbf{Eval} & \textbf{TB2.0} \\
            \textbf{Len} & \textbf{Len} & \textbf{YaRN2} & \textbf{YaRN2} & \\
            \midrule
            32,768 & 40,960 &            &            & \textbf{13.0 $\pm$ 2.2} \\
            32,768 & 65,536 &            & \checkmark & 11.9 $\pm$ 2.0 \\
            65,536 & 65,536 &            &            & 10.3 $\pm$ 2.0 \\
            65,536 & 65,536 & \checkmark & \checkmark & 11.9 $\pm$ 2.1 \\
            \bottomrule
        \end{tabular}}
    \end{minipage}
\end{table}

\subsection{Long Context Training and Evaluation}

Terminal trajectories can vary significantly in their turn counts (Appendix~\ref{appendix:synthetic-trajectory-analysis}), which in turn induces large differences in token counts depending on task type and difficulty. As shown in Appendix~\ref{appendix:synthetic-trajectory-analysis}, most trajectories fit within the default maximum sequence length of Qwen3 models (32,768 tokens), but a nontrivial subset exceeds this limit and is truncated by SFT. Motivated by this, we investigate long-context training for Qwen3-8B.

Specifically, we compare SFT with a 65,536-token context window without YaRN2 \citep{peng2023yarn}, SFT with YaRN2, and a baseline that only applies YaRN2 at evaluation time. Across these settings, we observe no significant performance differences (Table~\ref{tab:long_context}). Moreover, evaluating under the standard Qwen3-8B setup with a 40,960-token context window yields stronger results. Our results suggest that extending the context length slightly hurts performance; most high-quality supervision already fits within the standard window, while the long-tail trajectories tend to be noisy and less informative.



\subsection{Curriculum Learning}

\begin{figure}[h]
    \small
    \centering
    \begin{minipage}[c]{0.55\textwidth}
        \centering
        \includegraphics[width=\linewidth]{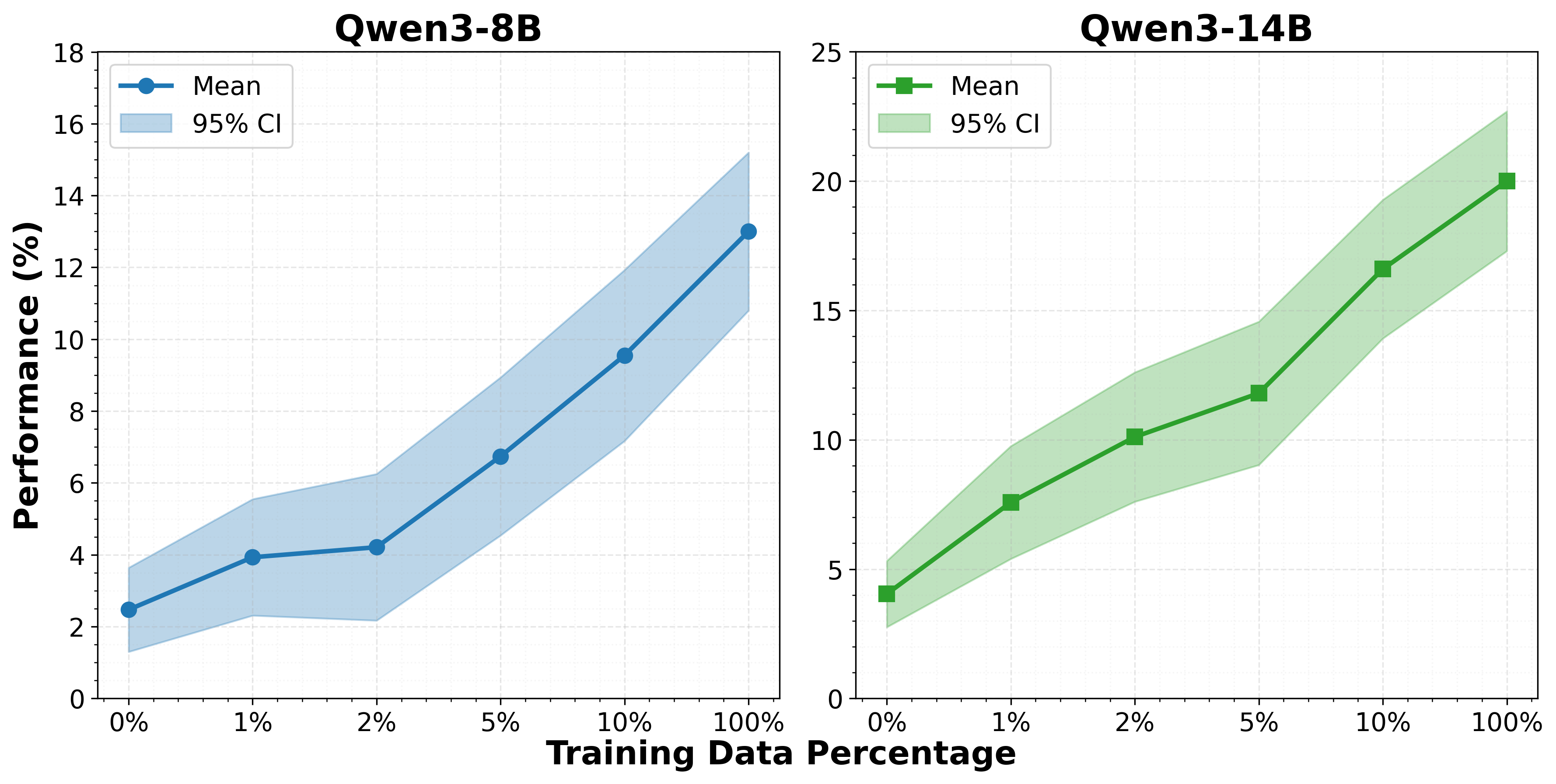}
        \caption{\textbf{Impact of training data scale on model performance.} Our scaling experiments show that TB2.0 performance increases with training data volume for both Qwen3-8B and Qwen3-14B.}
        \label{fig:scaling}
    \end{minipage}
    \hfill 
    \begin{minipage}[c]{0.4\textwidth}
        \centering
        \begin{tabular}{llc}
            \hline
            \textbf{Model} & \textbf{Strategy} & \textbf{TB2.0} \\
            \hline
            Qwen3-8B & mixed & \textbf{13.03 $\pm$ 2.16} \\
            Qwen3-8B & curriculum & 10.39 $\pm$ 1.71\\
            \hline
        \end{tabular}
        \vspace{10pt} 
        \captionof{table}{\textbf{Ablations on curriculum learning strategies.} We compare a mixed single-stage strategy with a two-stage curriculum. The mixed strategy achieves the strongest performance.}
        \label{tab:model_strategy_eval}
    \end{minipage}
\end{figure}

We investigate two SFT curriculum strategies for data mixing: (1) a two-stage curriculum, where we first train on dataset adapters, followed by synthetic task data; and (2) a single-stage strategy, where we train on all datasets concurrently. As demonstrated in Table~\ref{tab:model_strategy_eval}, the two-stage curriculum yields no performance advantage over simple mixed training. Consequently, we adopt the single-stage mixed training strategy for all other experiments in this paper.

\subsection{Scaling Experiments}
We investigate the impact of training data scale on model performance by fine-tuning Qwen3-8B and Qwen3-14B models on varying percentages of synthetic training data (0\%, 1\%, 2\%, 5\%, 10\% and 100\%). Both models demonstrate consistent performance improvements as training data increases (Figure~\ref{fig:scaling}). The larger 14B model not only achieves higher absolute performance across all data scales but also exhibits greater gains from additional training data. These results demonstrate that both model capacity and training data scale are critical factors for performance.

\section{Conclusion}

In this work, we address the data scarcity bottleneck in terminal agent training by introducing \textbf{\ourmethod}, a scalable framework that synergizes large-scale dataset adaptation with targeted synthetic task generation. Our systematic study demonstrates that precise data engineering enables the efficient \textbf{\ourmodel} family to significantly outperform its Qwen3 base and rival larger frontier models on Terminal-Bench 2.0, proving that high-quality, diverse trajectories are more pivotal than sheer parameter scale. Looking ahead, we see significant potential in extending this foundation with Reinforcement Learning (RL), leveraging verifiable execution feedback to enable self-correction and optimal planning for long-horizon tasks. We release our models and most of our synthetic datasets, including the adapter and skill-based task subsets, to democratize research in autonomous terminal agents.





\setcitestyle{numbers}
\bibliographystyle{plainnat}
\bibliography{main}

\clearpage
\appendix

\section{Appendix}

\begin{table*}[ht]
\centering
\small
\caption{Skills Summary by Domain}
\label{tab:skills_summary}
\begin{tabularx}{\linewidth}{l|X|X}
\toprule
\textbf{Domain} & \textbf{Skill Types} & \textbf{Example Skill} \\
\midrule
Security & Systems, Data Processing, Web Security, Algorithmic, Testing & Craft exploit payloads to bypass authentication and identify vulnerabilities \\
\hdashline
Software Engineering & Algorithmic, Systems, Data Processing, Web Security, Testing, Mathematical & Implement graph traversal (BFS/DFS) for dependency resolution \\
\hdashline
File Operations & File I/O, Navigation, Data Parsing, Transformation, Archives, Resources, Network & Parse structured formats (JSON/XML/CSV) with encoding and validation \\
\hdashline
Data Querying & Query Construction, Data Comprehension, Graph Processing, Result Processing & Writing queries using the formal syntax of declarative query languages for structured data \\
\hdashline
Data Science & Systems, Data Processing, Algorithmic, Mathematical, Testing, Web Security & Load and transform tabular data with groupby, filtering, and aggregation \\
\hdashline
Debugging & Systems, Debugging, Testing, Algorithmic, Data Processing, Mathematical & Resolve package dependency conflicts through constraint analysis \\
\hdashline
Scientific Computing & Data Processing, Algorithmic, Mathematical, Systems, Testing, Statistical, Web Security & Computing distance metrics between discrete probability distributions \\
\hdashline
Data Processing & Data I/O, Manipulation, String/Text, Algorithmic, Mathematical, Time Series, Systems, Testing & Build transformation pipelines with interpolation and feature extraction \\
\hdashline
System Administration & Filesystem, Process/Service, Network, Configuration, Deployment, Data Processing, Security, Shell Scripting, Testing, Algorithmic & Manage file permissions, configure services, and automate tasks with shell scripts \\
\bottomrule
\end{tabularx}
\end{table*}

\subsection{Synthetic Trajectory Analysis}
\label{appendix:synthetic-trajectory-analysis}

\begin{figure}[h]
    \centering 
    \includegraphics[width=0.88\columnwidth]{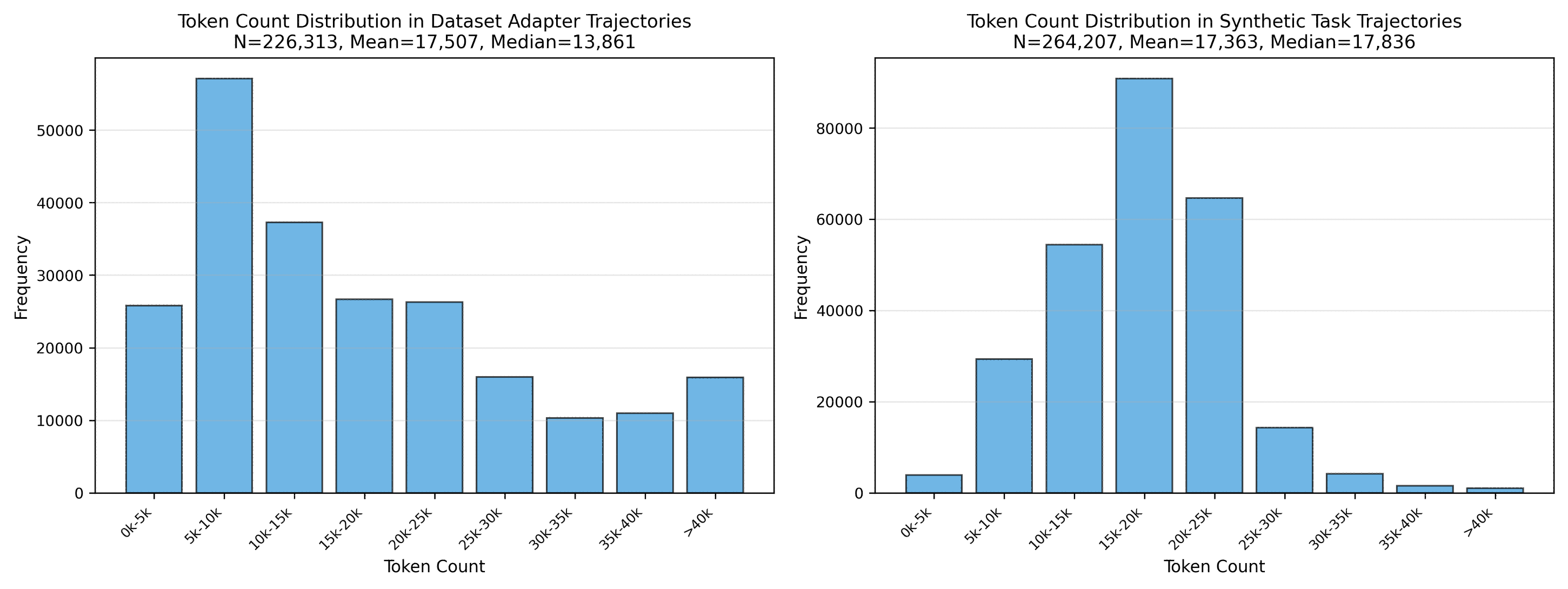}
    \caption{Distribution of \# tokens in the generated trajectories.}
    \label{fig:token_stats}
\end{figure}

\begin{figure}[h]
    \centering 
    \includegraphics[width=0.88\columnwidth]{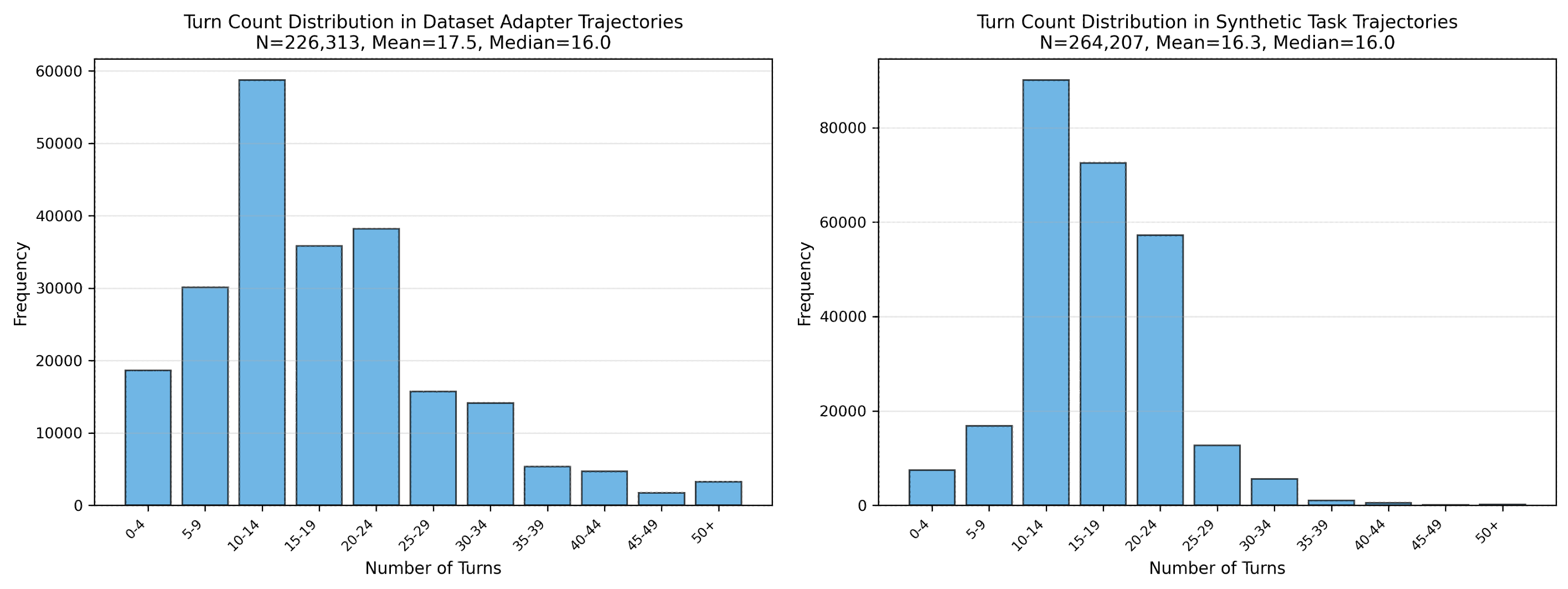}
    \caption{Distribution of \# turns in the generated trajectories.}
    \label{fig:turn_stats}
\end{figure}

We compute the distribution of the number of tokens (Figure~\ref{fig:token_stats}) and turns (Figure~\ref{fig:turn_stats}) in our trajectories, separating synthetic tasks from dataset adapters.

\subsection{Details for Trajectory Generation}
\label{appendix:trajectory-generation-templates}

We use Terminus 2 to generate all trajectories for SFT. We provide the full Terminus 2 system prompt template in Figure~\ref{tab:terminus_system_prompt}, which includes placeholders for \textit{instruction} and \textit{terminal\_state}. Our \ourmethod pipeline generates tasks with explicit instructions that replace the \textit{instruction} placeholder. For dataset adapters, we instead insert the original prompt followed by a domain-specific suffix, shown in Figures~\ref{tab:math_instruction_template}--\ref{tab:swe_instruction_template}. The \textit{terminal\_state} placeholder is populated with the latest terminal output summarizing the current shell state.


\begin{figure*}[h!] 
\centering
\begin{minipage}{0.98\textwidth} 
\centering
\begin{sectionbox}[]{Terminus 2 Agent: System Prompt Template}
    \centering
    \scriptsize 
    \begin{lstlisting}[
        basicstyle=\ttfamily\scriptsize, 
        breaklines=true, 
        columns=fullflexible, 
        xleftmargin=2pt, 
        xrightmargin=2pt,
        aboveskip=0pt, 
        belowskip=0pt
    ]
You are an AI assistant tasked with solving command-line tasks in a Linux environment. You will be given a task description and the output from previously executed commands. Your goal is to solve the task by providing batches of shell commands.

Format your response as JSON with the following structure:

{{
  "analysis": "Analyze the current state based on the terminal output provided. What do you see? What has been accomplished? What still needs to be done?",
  "plan": "Describe your plan for the next steps. What commands will you run and why? Be specific about what you expect each command to accomplish.",
  "commands": [
    {{
      "keystrokes": "ls -la\n",
      "duration": 0.1
    }},
    {{
      "keystrokes": "cd project\n",
      "duration": 0.1
    }}
  ],
  "task_complete": true
}}

Required fields:
- "analysis": Your analysis of the current situation
- "plan": Your plan for the next steps
- "commands": Array of command objects to execute

Optional fields:
- "task_complete": Boolean indicating if the task is complete (defaults to false if not present)

Command object structure:
- "keystrokes": String containing the exact keystrokes to send to the terminal (required)
- "duration": Number of seconds to wait for the command to complete before the next command will be executed (defaults to 1.0 if not present)

IMPORTANT: The text inside "keystrokes" will be used completely verbatim as keystrokes. Write commands exactly as you want them sent to the terminal:
- Most bash commands should end with a newline (\n) to cause them to execute
- For special key sequences, use tmux-style escape sequences:
  - C-c for Ctrl+C
  - C-d for Ctrl+D

The "duration" attribute specifies the number of seconds to wait for the command to complete (default: 1.0) before the next command will be executed. On immediate tasks (e.g., cd, ls, echo, cat) set a duration of 0.1 seconds. On commands (e.g., gcc, find, rustc) set a duration of 1.0 seconds. On slow commands (e.g., make, python3 [long running script], wget [file]) set an appropriate duration as you determine necessary.

It is better to set a smaller duration than a longer duration. It is always possible to wait again if the prior output has not finished, by running {{"keystrokes": "", "duration": 10.0}} on subsequent requests to wait longer. Never wait longer than 60 seconds; prefer to poll to see intermediate result status.

Important notes:
- Each command's keystrokes are sent exactly as written to the terminal
- Do not include extra whitespace before or after the keystrokes unless it's part of the intended command
- Extra text before or after the JSON will generate warnings but be tolerated
- The JSON must be valid - use proper escaping for quotes and special characters within strings
- Commands array can be empty if you want to wait without taking action

Task Description:
{instruction}

Current terminal state:
{terminal_state}
\end{lstlisting}
\end{sectionbox}
\caption{System prompt template for Terminus 2.}
\label{tab:terminus_system_prompt}
\end{minipage}
\end{figure*}


\begin{figure*}[h!]
\centering
\begin{minipage}{1.0\textwidth}
\vspace{0mm}
\centering
\begin{sectionbox}[]{Math Dataset Adapter: Instruction Template}
    \centering
    \footnotesize
    \begin{lstlisting}[
        basicstyle=\ttfamily\scriptsize, 
        breaklines=true, 
        columns=fullflexible, 
        xleftmargin=2pt, 
        xrightmargin=2pt,
        aboveskip=0pt, 
        belowskip=0pt
    ]
{math_prompt}

Please place your final answer in a file named `/app/solution.txt`.
    \end{lstlisting}
\end{sectionbox}
\vspace{-2mm}
\caption{Instruction template for math dataset adapter.}
\label{tab:math_instruction_template}
\end{minipage}
\end{figure*}


\begin{figure*}[h!]
\centering
\begin{minipage}{1.0\textwidth}
\vspace{0mm}
\centering
\begin{sectionbox}[]{Code Dataset Adapter: Instruction Template}
    \centering
    \footnotesize
    \begin{lstlisting}[
        basicstyle=\ttfamily\scriptsize, 
        breaklines=true, 
        columns=fullflexible, 
        xleftmargin=2pt, 
        xrightmargin=2pt,
        aboveskip=0pt, 
        belowskip=0pt
    ]
{code_prompt}

Write Python code to solve the problem. Please place the solution code in a file named `/app/solution.py`.
    \end{lstlisting}
\end{sectionbox}
\vspace{-2mm}
\caption{Instruction template for code dataset adapter.}
\label{tab:code_instruction_template}
\end{minipage}
\end{figure*}


\begin{figure*}[h!]
\centering
\begin{minipage}{1.0\textwidth}
\vspace{0mm}
\centering
\begin{sectionbox}[]{SWE Dataset Adapter: Instruction Template}
    \centering
    \footnotesize
    \begin{lstlisting}[
        basicstyle=\ttfamily\scriptsize, 
        breaklines=true, 
        columns=fullflexible, 
        xleftmargin=2pt, 
        xrightmargin=2pt,
        aboveskip=0pt, 
        belowskip=0pt
    ]
{swe_prompt}

Please first localize the bug based on the issue statement, generate *SEARCH/REPLACE* edits to fix the issue, and save the diff to a file named `/app/solution.patch`.
    \end{lstlisting}
\end{sectionbox}
\vspace{-2mm}
\caption{Instruction template for SWE dataset adapter.}
\label{tab:swe_instruction_template}
\end{minipage}
\end{figure*}

\subsection{Details for Synthetic Task Generation}
\paragraph{Curation of Primitive Skills} We showcase the skill types and examples for each domain in Table~\ref{tab:skills_summary}. These domains encompass a wide spectrum of capabilities required for robust terminal interaction. For instance, Security tasks involve skill types such as Systems, Data Processing, Web Security, Algorithmic, and Testing, with practical applications like crafting exploit payloads to bypass authentication. Similarly, Software Engineering combines Algorithmic and Systems skills to implement complex logic like graph traversals for dependency resolution. File Operations and System Administration focus on core infrastructure tasks, ranging from parsing structured formats (JSON/XML/CSV) to managing file permissions and automating service configurations. We also curate specialized skills for Data Science, Scientific Computing, and Data Processing, which require mathematical and statistical proficiency to build transformation pipelines or compute distance metrics between probability distributions. By systematically categorizing these primitive skills from low-level filesystem manipulation to high-level algorithmic reasoning, we ensure comprehensive coverage of the challenges an autonomous agent must navigate in a terminal environment.

\paragraph{Prompts for Synthetic Task Generation} We employ a modular prompting strategy that fuses a structural backbone with specialized constraints. Figure~\ref{tab:master_template} presents the system prompt, which governs general task logic, while Figures~\ref{tab:data_processing_module}--\ref{tab:software_engineering_module} detail the domain-specific requirement modules injected into this template. These modules define the preconditions and objectives for each vertical; for instance, the Security module mandates crafting exploit payloads to identify vulnerabilities, while the Data Processing module requires building transformation pipelines with interpolation. This ensures that generated tasks precisely align with the diverse primitive skills outlined previously.
\begin{figure*}[h!]
\centering
\begin{minipage}{1.0\textwidth}
\vspace{0mm}
\centering
\begin{sectionbox}[]{Skill-based Task Generation: Prompt Template}
    \centering
    \footnotesize
    \begin{tabular}{p{0.97\textwidth} c}
\textbf{SYSTEM PROMPT CONSTRUCTION:}\\
\\
\textbf{1. Role Definition:}\\
You are an expert at creating $\langle$domain$\rangle$ tasks for AI agent training.\\
\\
\textbf{2. Domain Context (Insert Variable Module Here):}\\
\textit{[INSERT CONTENT FROM SPECIFIC DOMAIN TABLES BELOW (e.g., Data Science, Security, etc.)]}\\
\\
\textbf{3. Universal Task Requirements:}\\
- \textbf{Challenging to solve:} Requires domain knowledge, analytical thinking, and efficient implementation.\\
- \textbf{Easy to verify:} Success must be determinable by programmatically checking outputs, exit codes, or system state.\\
- \textbf{Self-contained:} All necessary information must be in the prompt.\\
- \textbf{Realistic:} The problem should resemble tasks professionals face in this domain.\\
\\
\textbf{4. Output Format:}\\
You MUST output using these XML tags:\\
- $\langle$prompt$\rangle$: The task description with explicit requirements.\\
- $\langle$tests$\rangle$: Pytest functions to verify the solution.\\
- $\langle$weights$\rangle$: Test scoring distribution.\\
- $\langle$info$\rangle$: Task metadata.\\
- $\langle$files$\rangle$: Input data or initial file structure.\\
- $\langle$test\_requirements$\rangle$: Python packages required for testing.\\
\\
\textbf{5. Critical Rules:}\\
- \textbf{No Leakage:} Never include code that solves the task in the prompt.\\
- \textbf{Verification:} Prioritize tasks with clear, programmatic verification.\\
- \textbf{Originality:} Tasks should require thought, not just copying standard tutorials.\\
- \textbf{Complete Specification:} Include all information needed to complete the task (file paths, formats, constraints).\\
\\
\hline
\\
\textbf{USER MESSAGE CONSTRUCTION (Appended to System Prompt):}\\
\\
\# Task Generation Request\\
Category: $\langle$CATEGORY$\rangle$\\
\\
\#\# Primitive Skills (Building Blocks)\\
$\langle$Primitive\_Skills$\rangle$\\
\\
\#\# Pre-designed Docker Environment\\
Tasks will run in this pre-designed Docker environment:\\
$\langle$DOCKERFILE\_CONTENT$\rangle$\\
If additional packages are needed for testing, list them in $\langle$test\_requirements$\rangle$.
\\
\#\# Instructions\\
CREATE A NOVEL TASK that:\\
1. Combines 3-5 primitives in a creative, unexpected way\\
2. Is NOT a recreation of common coding challenges\\
3. Is challenging to solve but easy to verify\\
4. Has clear, unambiguous specifications\\
\\
Think of an original scenario or application - don't just combine primitives mechanically.
    \end{tabular}
\end{sectionbox}
\vspace{-2mm}
\caption{Prompt template used for all skill-based generation. Domain-specific modules are inserted into section 2.}
\label{tab:master_template}
\end{minipage}
\end{figure*}
\begin{figure*}[h!]
\centering
\begin{minipage}{1.0\textwidth}
\vspace{0mm}
\centering
\begin{sectionbox}[]{Domain Module: Data Processing}
    \centering
    \footnotesize
    \begin{tabular}{p{0.97\textwidth} c}
\textbf{\# Data Processing Task Builder}\\
You are an expert at creating data processing programming tasks for AI agent training.\\
\\
\textbf{Domain Focus}\\
Create tasks involving:\\
- **File format handling**: CSV, JSON, XML, Parquet, binary formats\\
- **Data transformation**: Cleaning, normalization, aggregation\\
- **ETL pipelines**: Extract, transform, load workflows\\
- **Stream processing**: Real-time data handling\\
- **Data validation**: Schema enforcement, error handling\\
\\
\textbf{Your Task}\\
Create a programming task that tests data processing skills. The task should be:\\
1. **Challenging to solve** - Requires domain knowledge and analytical thinking\\
2. **Easy to verify** - Success can be determined by checking outputs or state\\
3. **Self-contained** - All information needed is in the prompt\\
4. **Realistic** - Resembles tasks a professional might actually face
    \end{tabular}
\end{sectionbox}
\vspace{-2mm}
\caption{Module for Data Processing tasks.}
\label{tab:data_processing_module}
\end{minipage}
\end{figure*}

\begin{figure*}[h!]
\centering
\begin{minipage}{1.0\textwidth}
\vspace{0mm}
\centering
\begin{sectionbox}[]{Domain Module: Data Querying}
    \centering
    \footnotesize
    \begin{tabular}{p{0.97\textwidth} c}
\textbf{\# Data Querying Task Builder}\\
You are an expert at creating data querying programming tasks for AI agent training.\\
\\
\textbf{Domain Focus}\\
Create tasks involving:\\
- **SQL operations**: Complex joins, window functions, CTEs\\
- **Query optimization**: Indexes, execution plans, performance\\
- **Database operations**: Schema design, migrations, constraints\\
- **NoSQL patterns**: Document, key-value, graph queries\\
- **Data retrieval**: Pagination, filtering, full-text search\\
\\
\textbf{Your Task}\\
Create a programming task that tests data querying skills. The task should be:\\
1. **Challenging to solve** - Requires domain knowledge and analytical thinking\\
2. **Easy to verify** - Success can be determined by checking outputs or state\\
3. **Self-contained** - All information needed is in the prompt\\
4. **Realistic** - Resembles tasks a professional might actually face
    \end{tabular}
\end{sectionbox}
\vspace{-2mm}
\caption{Module for Data Querying tasks.}
\label{tab:data_querying_module}
\end{minipage}
\end{figure*}

\begin{figure*}[h!]
\centering
\begin{minipage}{1.0\textwidth}
\vspace{0mm}
\centering
\begin{sectionbox}[]{Domain Module: Data Science}
    \centering
    \footnotesize
    \begin{tabular}{p{0.97\textwidth} c}
\textbf{\# Data Science Task Builder}\\
You are an expert at creating data science programming tasks for AI agent training.\\
\\
\textbf{Domain Focus}\\
Create tasks involving:\\
- **Exploratory Analysis**: Statistical summaries, visualization, pattern discovery\\
- **Feature Engineering**: Transformation, encoding, selection, creation\\
- **Statistical Modeling**: Regression, hypothesis testing, Bayesian analysis\\
- **Data Mining**: Clustering, association rules, anomaly detection\\
- **Reporting**: Automated insights, metric computation, summary generation\\
\\
\textbf{Your Task}\\
Create a programming task that tests data science skills. The task should be:\\
1. **Challenging to solve** - Requires statistical thinking and data intuition\\
2. **Easy to verify** - Success can be determined by checking outputs or metrics\\
3. **Self-contained** - All information needed is in the prompt\\
4. **Realistic** - Resembles tasks a data scientist might actually face
    \end{tabular}
\end{sectionbox}
\vspace{-2mm}
\caption{Module for Data Science tasks.}
\label{tab:data_science_module}
\end{minipage}
\end{figure*}

\begin{figure*}[h!]
\centering
\begin{minipage}{1.0\textwidth}
\vspace{0mm}
\centering
\begin{sectionbox}[]{Domain Module: Debugging}
    \centering
    \footnotesize
    \begin{tabular}{p{0.97\textwidth} c}
\textbf{\# Debugging Task Builder}\\
You are an expert at creating debugging programming tasks for AI agent training.\\
\\
\textbf{Domain Focus}\\
Create tasks involving:\\
- **Error diagnosis**: Stack traces, logs, error messages\\
- **Root cause analysis**: Bisection, delta debugging\\
- **Performance debugging**: Profiling, bottleneck identification\\
- **Memory issues**: Leaks, corruption, allocation problems\\
- **Concurrency bugs**: Race conditions, deadlocks, livelocks\\
\\
\textbf{Your Task}\\
Create a programming task that tests debugging skills. The task should be:\\
1. **Challenging to solve** - Requires domain knowledge and analytical thinking\\
2. **Easy to verify** - Success can be determined by checking outputs or state\\
3. **Self-contained** - All information needed is in the prompt\\
4. **Realistic** - Resembles tasks a professional might actually face
    \end{tabular}
\end{sectionbox}
\vspace{-2mm}
\caption{Module for Debugging tasks.}
\label{tab:debugging_module}
\end{minipage}
\end{figure*}

\begin{figure*}[h!]
\centering
\begin{minipage}{1.0\textwidth}
\vspace{0mm}
\centering
\begin{sectionbox}[]{Domain Module: File Operations}
    \centering
    \footnotesize
    \begin{tabular}{p{0.97\textwidth} c}
\textbf{\# File Operations Task Builder}\\
You are an expert at creating file operations programming tasks for AI agent training.\\
\\
\textbf{Domain Focus}\\
Create tasks involving:\\
- **File I/O**: Reading, writing, appending, seeking\\
- **Directory operations**: Traversal, creation, permissions\\
- **File formats**: Binary, text, structured data\\
- **Compression**: Zip, tar, gzip, custom formats\\
- **File system operations**: Links, permissions, metadata\\
\\
\textbf{Your Task}\\
Create a programming task that tests file operations skills. The task should be:\\
1. **Challenging to solve** - Requires domain knowledge and analytical thinking\\
2. **Easy to verify** - Success can be determined by checking outputs or state\\
3. **Self-contained** - All information needed is in the prompt\\
4. **Realistic** - Resembles tasks a professional might actually face
    \end{tabular}
\end{sectionbox}
\vspace{-2mm}
\caption{Module for File Operations tasks.}
\label{tab:file_operations_module}
\end{minipage}
\end{figure*}

\begin{figure*}[h!]
\centering
\begin{minipage}{1.0\textwidth}
\vspace{0mm}
\centering
\begin{sectionbox}[]{Domain Module: Scientific Computing}
    \centering
    \footnotesize
    \begin{tabular}{p{0.97\textwidth} c}
\textbf{\# Scientific Computing Task Builder}\\
You are an expert at creating scientific computing programming tasks for AI agent training.\\
\\
\textbf{Domain Focus}\\
Create tasks involving:\\
- **Numerical simulation**: ODEs, PDEs, Monte Carlo\\
- **Signal processing**: FFT, filtering, spectral analysis\\
- **Statistical analysis**: Hypothesis testing, regression, sampling\\
- **Visualization**: Plotting, data exploration\\
- **Domain-specific**: Physics, biology, chemistry applications\\
\\
\textbf{Your Task}\\
Create a programming task that tests scientific computing skills. The task should be:\\
1. **Challenging to solve** - Requires domain knowledge and analytical thinking\\
2. **Easy to verify** - Success can be determined by checking outputs or state\\
3. **Self-contained** - All information needed is in the prompt\\
4. **Realistic** - Resembles tasks a professional might actually face
    \end{tabular}
\end{sectionbox}
\vspace{-2mm}
\caption{Module for Scientific Computing tasks.}
\label{tab:scientific_computing_module}
\end{minipage}
\end{figure*}

\begin{figure*}[h!]
\centering
\begin{minipage}{1.0\textwidth}
\vspace{0mm}
\centering
\begin{sectionbox}[]{Domain Module: Security}
    \centering
    \footnotesize
    \begin{tabular}{p{0.97\textwidth} c}
\textbf{\# Security Task Builder}\\
You are an expert at creating security programming tasks for AI agent training.\\
\\
\textbf{Domain Focus}\\
Create tasks involving:\\
- **Cryptography**: Encryption, decryption, key management, hash functions\\
- **Vulnerability Analysis**: Code review, exploit identification, security auditing\\
- **Authentication**: Password handling, token validation, session management\\
- **Network Security**: Protocol analysis, traffic inspection, firewall rules\\
- **Secure Coding**: Input validation, output encoding, secure defaults\\
\\
\textbf{Your Task}\\
Create a programming task that tests security skills. The task should be:\\
1. **Challenging to solve** - Requires security knowledge and analytical thinking\\
2. **Easy to verify** - Success can be determined by checking outputs or state\\
3. **Self-contained** - All information needed is in the prompt\\
4. **Realistic** - Resembles tasks a security engineer might actually face
    \end{tabular}
\end{sectionbox}
\vspace{-2mm}
\caption{Module for Security tasks.}
\label{tab:security_module}
\end{minipage}
\end{figure*}

\begin{figure*}[h!]
\centering
\begin{minipage}{1.0\textwidth}
\vspace{0mm}
\centering
\begin{sectionbox}[]{Domain Module: Software Engineering}
    \centering
    \footnotesize
    \begin{tabular}{p{0.97\textwidth} c}
\textbf{\# Software Engineering Task Builder}\\
You are an expert at creating software engineering programming tasks for AI agent training.\\
\\
\textbf{Domain Focus}\\
Create tasks involving:\\
- **Code quality**: Refactoring, testing, documentation\\
- **Build systems**: Compilation, linking, packaging\\
- **Version control**: Git operations, merge conflicts\\
- **API design**: REST, GraphQL, protocol design\\
- **Architecture**: Patterns, modularity, scalability\\
\\
\textbf{Your Task}\\
Create a programming task that tests software engineering skills. The task should be:\\
1. **Challenging to solve** - Requires domain knowledge and analytical thinking\\
2. **Easy to verify** - Success can be determined by checking outputs or state\\
3. **Self-contained** - All information needed is in the prompt\\
4. **Realistic** - Resembles tasks a professional might actually face
    \end{tabular}
\end{sectionbox}
\vspace{-2mm}
\caption{Module for Software Engineering tasks.}
\label{tab:software_engineering_module}
\end{minipage}
\end{figure*}

\begin{figure*}[h!]
\centering
\begin{minipage}{1.0\textwidth}
\vspace{0mm}
\centering
\begin{sectionbox}[]{Domain Module: System Administration}
    \centering
    \footnotesize
    \begin{tabular}{p{0.97\textwidth} c}
\textbf{\# System Administration Task Builder}\\
You are an expert at creating system administration programming tasks for AI agent training.\\
\\
\textbf{Domain Focus}\\
Create tasks involving:\\
- **Process management**: Services, daemons, scheduling\\
- **Network configuration**: Routing, firewall, DNS\\
- **Storage management**: Filesystems, RAID, backups\\
- **Monitoring**: Logging, alerting, metrics\\
- **Automation**: Scripts, configuration management\\
\\
\textbf{Your Task}\\
Create a programming task that tests system administration skills. The task should be:\\
1. **Challenging to solve** - Requires domain knowledge and analytical thinking\\
2. **Easy to verify** - Success can be determined by checking outputs or state\\
3. **Self-contained** - All information needed is in the prompt\\
4. **Realistic** - Resembles tasks a professional might actually face
    \end{tabular}
\end{sectionbox}
\vspace{-2mm}
\caption{Module for System Administration tasks.}
\label{tab:system_administration_module}
\end{minipage}
\end{figure*}

\end{document}